%% file: main.tex
\PassOptionsToPackage{table,xcdraw}{xcolor}
\documentclass[sigconf]{acmart}
\usepackage{xcolor} 
\usepackage{booktabs}              
\usepackage{caption}              
\usepackage{graphicx}              
\usepackage{multirow} 
\usepackage{float}

\AtBeginDocument{%
  }

\copyrightyear{2025}
\acmYear{2025}
\setcopyright{none}
\acmConference[MM '25]{Proceedings of the 33rd ACM International Conference on Multimedia}{October 27--31, 2025}{Dublin, Ireland}
\acmBooktitle{Proceedings of the 33rd ACM International Conference on Multimedia (MM '25), October 27--31, 2025, Dublin, Ireland}
\acmDOI{10.1145/3746027.3754798}
\acmISBN{979-8-4007-2035-2/2025/10}

\acmSubmissionID{6726}

\begin{document}

\title[RAIDX: A RAG-GRPO Framework for Explainable Deepfake Detection]{RAIDX: A Retrieval-Augmented Generation and GRPO Reinforcement Learning Framework for Explainable Deepfake Detection}

\author{Tianxiao Li}
\orcid{0009-0001-8315-2135}
\affiliation{%
  \institution{University of Liverpool}
  \city{Liverpool}
  \country{United Kingdom}
}
\email{sgtli18@liverpool.ac.uk}

\author{Zhenglin Huang}
\orcid{0009-0005-3759-7888}
\affiliation{%
  \institution{University of Liverpool}
  \city{Liverpool}
  \country{United Kingdom}
}
\email{zhenglin@liverpool.ac.uk}

\author{Haiquan Wen}
\orcid{0009-0009-3804-6753}
\affiliation{%
  \institution{University of Liverpool}
  \city{Liverpool}
  \country{United Kingdom}
}
\email{sghwen4@liverpool.ac.uk}

\author{Yiwei He}
\affiliation{
  \institution{University of Liverpool}
  \city{Liverpool}
  \country{United Kingdom}
}
\email{yiwei.he@liverpool.ac.uk}

\author{Shuchang Lyu}
\affiliation{
  \institution{Beihang University}
  \city{Beijing}
  \country{China}
}
\email{lyushuchang@buaa.edu.cn}

\author{Baoyuan Wu}
\affiliation{
  \institution{The Chinese University of Hong Kong}
  \city{Shenzhen}
  \country{China}
}
\email{wubaoyuan1987@gmail.com}

\author{Guangliang Cheng}
\authornote{Corresponding author.}
\affiliation{
  \institution{University of Liverpool}
  \city{Liverpool}
  \country{United Kingdom}
}
\email{guangliang.cheng@liverpool.ac.uk}

\renewcommand{\shortauthors}{Li et al.}

\input{sec/0_abstract}

\maketitle

\input{sec/1_intro}
\input{sec/2_related_work}
\input{sec/3_method}
\input{sec/4_experiments}

\input{sec/5_conclusion}

\newpage

\begin{acks}
This work is supported by The Alan Turing Institute (UK) through the project \textit{Turing-DSO Labs Singapore Collaboration} (SDCfP2\textbackslash100009).
\end{acks}

{
    \small
    \bibliographystyle{ACM-Reference-Format}
    \bibliography{main}
}

\end{document}

%% file: sec/0_abstract.tex
\begin{abstract}
\label{sec:abstract}

The rapid advancement of AI-generation models has enabled the creation of hyperrealistic imagery, posing ethical risks through widespread misinformation. Current deepfake detection methods, categorized as face-specific detectors or general AI-generated detectors, lack transparency by framing detection as a classification task without explaining decisions. While several LLM-based approaches offer explainability, they suffer from coarse-grained analyses and dependency on labor-intensive annotations. This paper introduces \textbf{RAIDX} 
(\textbf{R}etrieval-\textbf{A}ugmented \textbf{I}mage \textbf{D}eepfake Detection and E\textbf{X}plainability), a novel deepfake detection framework integrating Retrieval-Augmented Generation (RAG) and Group Relative Policy Optimization (GRPO) to enhance detection accuracy and decision explainability. Specifically, RAIDX leverages RAG to incorporate external knowledge for improved detection accuracy and employs GRPO to autonomously generate fine-grained textual explanations and saliency maps, eliminating the need for extensive manual annotations. Experiments on multiple benchmarks demonstrate RAIDX’s effectiveness in identifying real or fake, and providing interpretable rationales in both textual descriptions and saliency maps, achieving state-of-the-art detection performance while advancing transparency in deepfake identification. RAIDX represents the first unified framework to synergize RAG and GRPO, addressing critical gaps in accuracy and explainability. Our code and models will be publicly available.
\end{abstract}

\ccsdesc[500]{Computing methodologies~Computer vision}
\ccsdesc[500]{Computing methodologies~Reinforcement learning}
\ccsdesc[300]{Security and privacy~Human and societal aspects of security and privacy}
\ccsdesc[100]{Computing methodologies~Natural language generation}
\ccsdesc[100]{Human-centered computing~Heat maps}

\keywords{deepfake detection, vision-language model, retrieval-augmented generation, reinforcement learning, multimodal explainability}

%% file: sec/1_intro.tex
\section{Introduction}
\label{sec:intro}

\begin{figure}[ht]
  \includegraphics[width=0.48\textwidth]{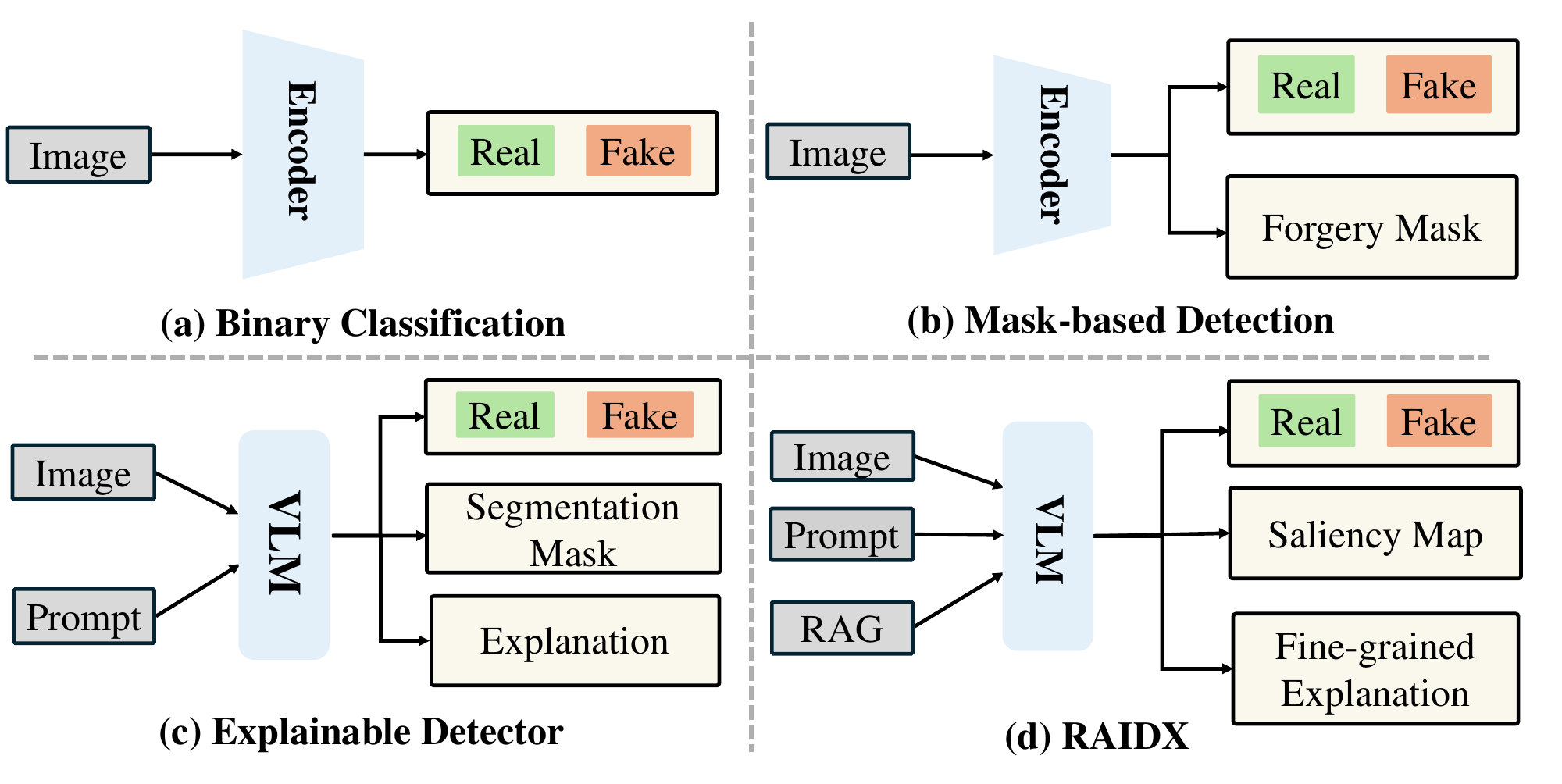}
  \caption{The RAIDX achieves detection, localization, and GRPO-enhanced explanation, all without requiring any supervision.}
  \label{fig:teaser}
\end{figure}

The rapid evolution of AI-generated visual content methods~\cite{DBLP:journals/corr/abs-2406-14555, DBLP:journals/pami/ZhanYWZLLKTX23} has positioned this field as one of the most transformative frontiers in computer vision. Driven by breakthroughs in generative modeling techniques—such as Generative Adversarial Networks (GANs)~\cite{goodfellow2014generativeadversarialnetworks,DBLP:journals/pami/KarrasLA21}, diffusion models~\cite{DBLP:conf/cvpr/RombachBLEO22, DBLP:journals/pami/CroitoruHIS23}, and Vision-Language Models (VLMs)~\cite{liu2023visual, Radford2018ImprovingLU, DBLP:journals/corr/abs-2402-03300}—these technologies now produce hyperrealistic imagery indistinguishable from authentic photographs. While such advancements unlock creative potential, they also introduce profound ethical risks, enabling malicious actors to organize misinformation campaigns and destabilize trust in digital media. This duality underscores an urgent need for robust and explainable deepfake detection systems capable of distinguishing synthetic content from genuine imagery, as well as explaining their judgment basis.

Current detection methodologies primarily fall into two categories: face-specific detectors~\cite{DBLP:conf/nips/YanYCZFZLWDWY24, kundu2024towards,zheng2021exploring}, which target facial manipulation artifacts, and general AI-generated content detectors~\cite{duan2024test,tan2024rethinking,tan2023learning}, designed for broader synthetic media identification. Despite progress, these approaches predominantly frame detection as a binary or multi-category classification task (shown in Figure~\ref{fig:teaser} (a)), offering {\it no transparency} into the rationale behind their decisions. This black-box nature limits practical utility, as users cannot verify why an image is flagged as fake. To address this, some mask-based detection methods~\cite{DBLP:journals/pami/DongCHCL23, DBLP:conf/cvpr/GuoLRGM023, DBLP:journals/tcsv/LiuLCL22} attempt to simultaneously provide corresponding mask predictions while performing classification (shown in Figure~\ref{fig:teaser} (b)). In recent years, with the rapid development of large language models (LLMs), although numerous LLM-based approaches ~\cite{kang2025legionlearninggroundexplain, huang2025sidasocialmediaimage} strive to provide explainability~(shown in Figure~\ref{fig:teaser} (c)), they remain constrained by two fundamental limitations: (1) \textbf{coarse-grained analyses that fail to pinpoint specific manipulation indicators}, and (2) \textbf{dependency on labor-intensive annotations} (e.g., mask labeling or textual descriptions), hindering real-world applicability.

Addressing these limitations requires rethinking both \textbf{how} explanations are generated and \textbf{what} evidence underpins detection. Innovations in LLM optimization, particularly reinforcement learning techniques like GRPO~\cite{deepseekai2024deepseekv3technicalreport}, offer a promising pathway. GRPO enhances reasoning by optimizing grouped outputs without auxiliary critic models, demonstrating success in diverse domains~\cite{huang2025visionr1incentivizingreasoningcapability, DBLP:journals/corr/abs-2502-14669}. Yet, its potential for deepfake detection remains unexplored. Additionally, in LLM-based approaches, RAG effectively integrates external knowledge to boost accuracy and produce precise text descriptions. Yet, its potential in VLMs or deepfake detection task remains underexplored.

We propose the \textbf{R}etrieval-\textbf{A}ugmented \textbf{I}mage \textbf{D}eepfake Detection and E\textbf{X}plainability (\textbf{RAIDX}), a novel framework that synergizes Retrieval-Augmented Generation RAG~\cite{DBLP:conf/nips/LewisPPPKGKLYR020} with GRPO to achieve state-of-the-art detection accuracy while producing interpretable, fine-grained explanations (shown in Figure~\ref{fig:teaser} (d)). Specifically,  RAIDX utilizes RAG~\cite{DBLP:conf/nips/LewisPPPKGKLYR020} to establish a retrieval repository to boost detection accuracy, while implementing GRPO to uncover interpretable rationales for model decisions, eliminating reliance on laborious manual annotation. To our knowledge, RAIDX represents the first framework to integrate both RAG and GRPO for deepfake detection, achieving enhanced detection accuracy while providing fine-grained explanations of its decision-making process.

The main contributions of this paper are as follows: 
\begin{itemize}
\item We introduce RAIDX, a novel deepfake detection framework to harness both RAG and the GRPO algorithm to enhance both detection accuracy and decision explainability.
\item We are the first to integrate RAG into deepfake detection, enhancing detection accuracy by incorporating external knowledge.
\item A novel GRPO-driven reinforcement learning framework is proposed to automatically generate fine-grained textual descriptions and precise saliency maps, localizing suspected fake regions and providing interpretable explanations without manual annotations. 
\item Extensive experiments on multiple deepfake detection benchmarks demonstrate that RAIDX effectively identifies and delineates suspected fake regions within images.
\end{itemize}

\begin{figure*}[t]
  \centering
  \includegraphics[width=0.9\textwidth]{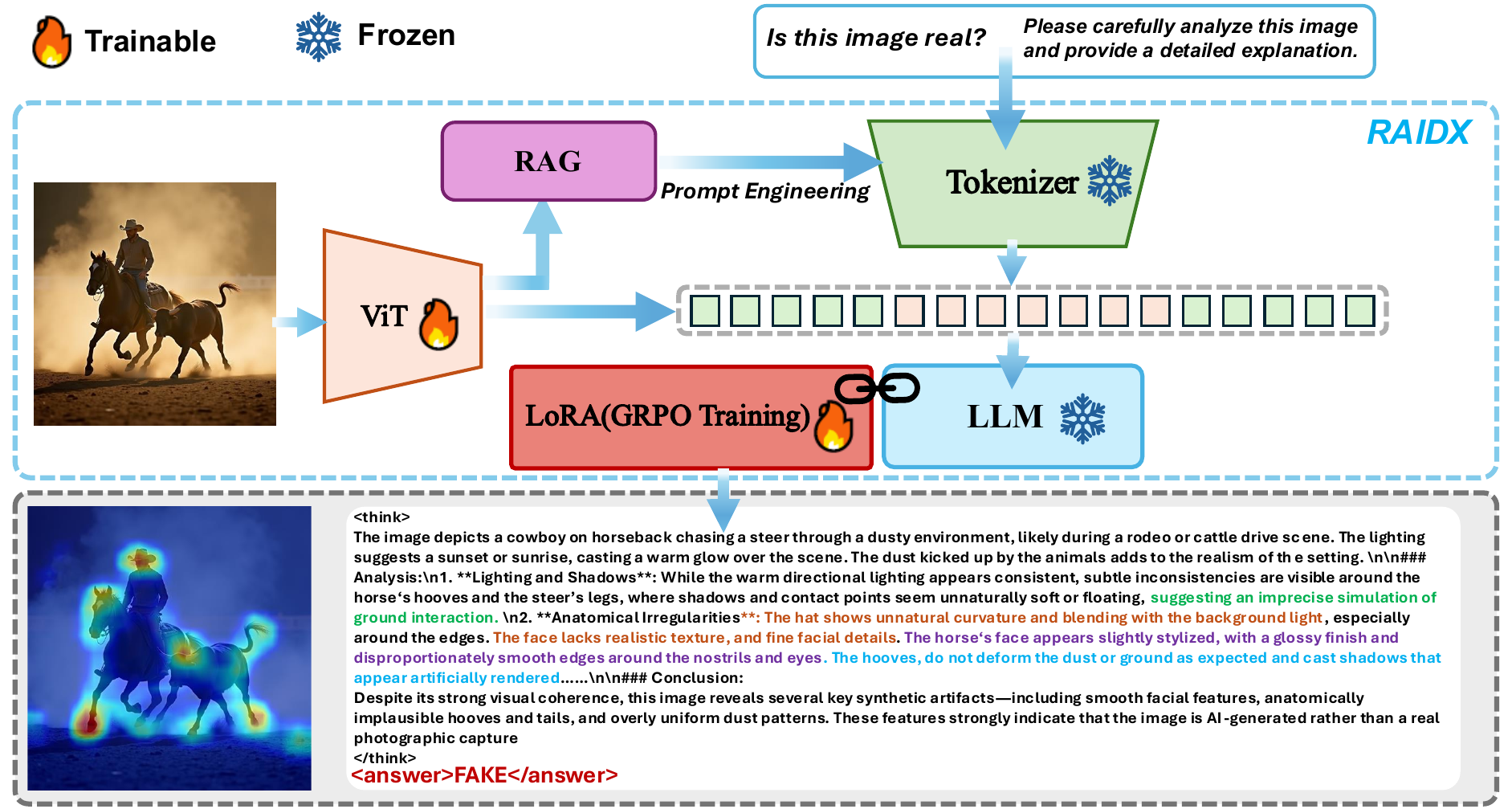}
  \caption{Framework of RAIDX: It integrates retrieval-augmented prompting with GRPO-enhanced reasoning for deepfake detection. A ViT and RAG module construct informative prompts using similar images, which are processed by a partially trainable LLM to output saliency-guided localization and fine-grained explanations without requiring any supervision.}
  \label{fig:overview}
\end{figure*}

%% file: sec/2_related_work.tex
\section{Related Work}
\label{sec:related}

\subsection{Vision-Language Models}
Vision-language models (VLMs) have emerged as powerful tools for multimodal tasks, including deepfake detection. The rapid advancement of large language models (LLMs) such as GPT-4 \cite{achiam2023gpt} and LLaMA \cite{touvron2023llamaopenefficientfoundation} has driven the evolution of VLMs, enabling seamless integration of vision and language modalities. Early approaches to VLMs, such as CLIP \cite{radford2021learning}, utilized contrastive learning to align vision and language representations in a shared embedding space. Subsequent models like BLIP \cite{li2022blip} introduced more advanced pretraining strategies, incorporating retrieval-augmented learning and cross-attention mechanisms to enhance contextual understanding.

Recent advancements, such as LLaVA \cite{liu2023visual} and Qwen2-VL \cite{wang2024qwen2vlenhancingvisionlanguagemodels}, have integrated ViT with LLMs, enabling vision-conditioned text generation and multimodal reasoning. Building on these foundations, models like Qwen2.5-Omni \cite{xu2025qwen25omnitechnicalreport} introduce novel architectures and techniques to handle complex multimodal tasks, such as video dialogue and reasoning, while maintaining strong performance in visual and linguistic understanding.

\subsection{Deepfake Detection Methods}
Image deepfake detection has evolved significantly with advancements in AI technologies. Traditional methods \cite{Cao_2022_CVPR_end2end, chen2022selfsupervisedlearningadversarialexample, wang2021representativeforgeryminingfake}, primarily based on Convolutional Neural Networks (CNNs), focus on identifying pixel-level anomalies such as unnatural textures and lighting inconsistencies. These Non-VLM-based approaches, exemplified by models like XceptionNet \cite{chollet2017xceptiondeeplearningdepthwise} and MesoNet \cite{MesoNet_2018}, have shown high precision in specific scenarios but often struggle with generalization due to their reliance on dataset-specific features.

More recently, the integration of VLMs has introduced a new paradigm in deepfake detection. VLM-based methods leverage the reasoning and cross-modal capabilities of these models to provide more flexible and interpretable detection. Models such as AntifakePrompt \cite{DBLP:journals/corr/abs-2310-17419}, ForgeryGPT \cite{liu2025forgerygptmultimodallargelanguage}, ForgeryTalker \cite{lian2024largescaleinterpretablemultimodalitybenchmark}, ForgerySleuth \cite{sun2024forgerysleuthempoweringmultimodallarge}, SIDA \cite{huang2025sidasocialmediaimage} and So-Fake-R1 \cite{huang2025sofakebenchmarkingexplainingsocial} utilize the capabilities of VLM to enhance detection accuracy and provide robust support for localization and explanation tasks. 
In the video domain, BusterX \cite{wen2025busterxmllmpoweredaigeneratedvideo} and BusterX++ \cite{wen2025busterxunifiedcrossmodalaigenerated} are outstanding representatives.
Most of these methods rely heavily on extensive manual annotation of masks or textual explanatory descriptions. Additionally, the interpretable descriptions learned by current models tend to be coarse. In contrast, RAIDX relies on the GRPO reinforcement learning strategy, which can produce relatively accurate saliency maps and fine-grained textual explanation outputs without requiring any manual text or mask annotations.

\subsection{Retrieval-Augmented Generation}
Retrieval-Augmented Generation (RAG) \cite{gao2024retrievalaugmentedgenerationlargelanguage} is a general paradigm which enhances LLMs by including relevant information retrieved from external databases into the input. RAG typically consists of three phases: \textbf{indexing}, \textbf{retrieval}, and \textbf{generation}. \textbf{Indexing} constructs external databases and their retrieval index from external data sources. \textbf{Retrieval} utilizes these indexes to fetch the relevant document chunks as context, given a user query. \textbf{Generation} integrates the retrieved context into the input prompt for LLMs, and LLMs then generate the final output based on the augmented inputs. RAG has been widely used in various domains \cite{shi2022raceretrievalaugmentedcommitmessage, zhang2023repocoderrepositorylevelcodecompletion, lewis2021retrievalaugmentedgenerationknowledgeintensivenlp}. Although RAG has achieved many successful cases in LLM tasks, there are few examples \cite{DBLP:journals/corr/abs-2410-08876, DBLP:journals/corr/abs-2411-13093} of its integration with VLMs. To the best of our knowledge, RAIDX is the first to apply RAG in the deepfake detection task.

\subsection{Reinforcement Learning in VLMs}
The integration of reinforcement learning (RL) into multimodal systems has emerged as a promising direction, particularly for tasks requiring alignment between vision and language. Recent advances in techniques such as Proximal Policy Optimization (PPO) \cite{schulman2017proximal} and Group Relative Policy Optimization (GRPO) \cite{deepseekai2025deepseekr1incentivizingreasoningcapability} have demonstrated their potential to enhance reasoning quality and factual correctness in language models, often by optimizing model outputs through structured feedback. GRPO, in particular, has gained significant attention due to its effectiveness in improving the reasoning capabilities of LLMs by leveraging rule-based rewards. In the visual domain, the application of RL to VLMs has shown substantial promise \cite{shen2025vlmr1, chen2025r1v, huang2025visionr1incentivizingreasoningcapability}. Experimental results from the VLM-R1 \cite{shen2025vlmr1} demonstrate that RL-based models not only deliver competitive performance on visual understanding tasks but also surpass supervised fine-tuning (SFT) in generalization ability. To the best of our knowledge, RAIDX is the first to apply the GRPO reinforcement learning training strategy to the field of deepfake detection, which significantly reduces reliance on manual mask annotations and text annotations—forming the foundation of our annotation-efficient approach.

%% file: sec/3_method.tex
\section{Method}
\label{sec:method}

In this section, we begin by outlining the overall architecture of our proposed RAIDX framework in Section~\ref{architecture}. This framework seamlessly integrates a vision encoder, a Retrieval-Augmented Generation (RAG) module, and a partially trainable large language model to enable multimodal deepfake detection and explanation. Next, in Section~\ref{training}, we elaborate on the GRPO-based reinforcement learning approach used for tuning LoRA~\cite{hu2022lora} adapters, enhancing both interpretability and robustness without the need for pixel-level annotations or manually crafted explanation labels.

As shown in Figure~\ref{fig:overview}, a central innovation of RAIDX is its seamless integration of Retrieval-Augmented Generation (RAG) and Group Relative Policy Optimization (GRPO), which together enable precise and interpretable reasoning. The RAG module bolsters factual accuracy by incorporating example-driven knowledge into the prompt, while GRPO refines a lightweight LoRA explanation head to enhance reasoning consistency and detection accuracy. This training approach empowers the model to self-improve via structured feedback and naturally develop Chain-of-Thought style reasoning capabilities.

\subsection{Architecture Design}
\label{architecture}

We introduce RAIDX, a Retrieval-Augmented framework engineered for multimodal deepfake detection and explainability. RAIDX integrates four core components: a Vision Transformer (ViT) for image feature extraction, a frozen tokenizer for encoding textual instructions, a Retrieval-Augmented Generation (RAG) module for knowledge grounding via exemplar-based reasoning, and a partially trainable Large Language Model (LLM) augmented with LoRA adapters, optimized through GRPO reinforcement learning. The complete framework is illustrated in Figure~\ref{fig:overview}. Each module is described in detail below:

\noindent
\textbf{Vision Transformer.}
The Vision Transformer (ViT) serves as the visual encoder to extract high-level visual features from input images. These extracted features are used in two parallel pathways: one is fed into the RAG module for FAISS-based retrieval, and the other is fused with textual tokens and processed jointly by the language model.

\noindent
\textbf{Tokenizer.}
The text tokenizer processes input prompts by converting them into a structured format. It combines user prompts with the RAG module's output to form a cohesive prompt sequence. An example of a user input prompt is shown in Figure~\ref{fig:overview}.

\begin{figure}[t]  
  \centering
  \includegraphics[width=\linewidth]{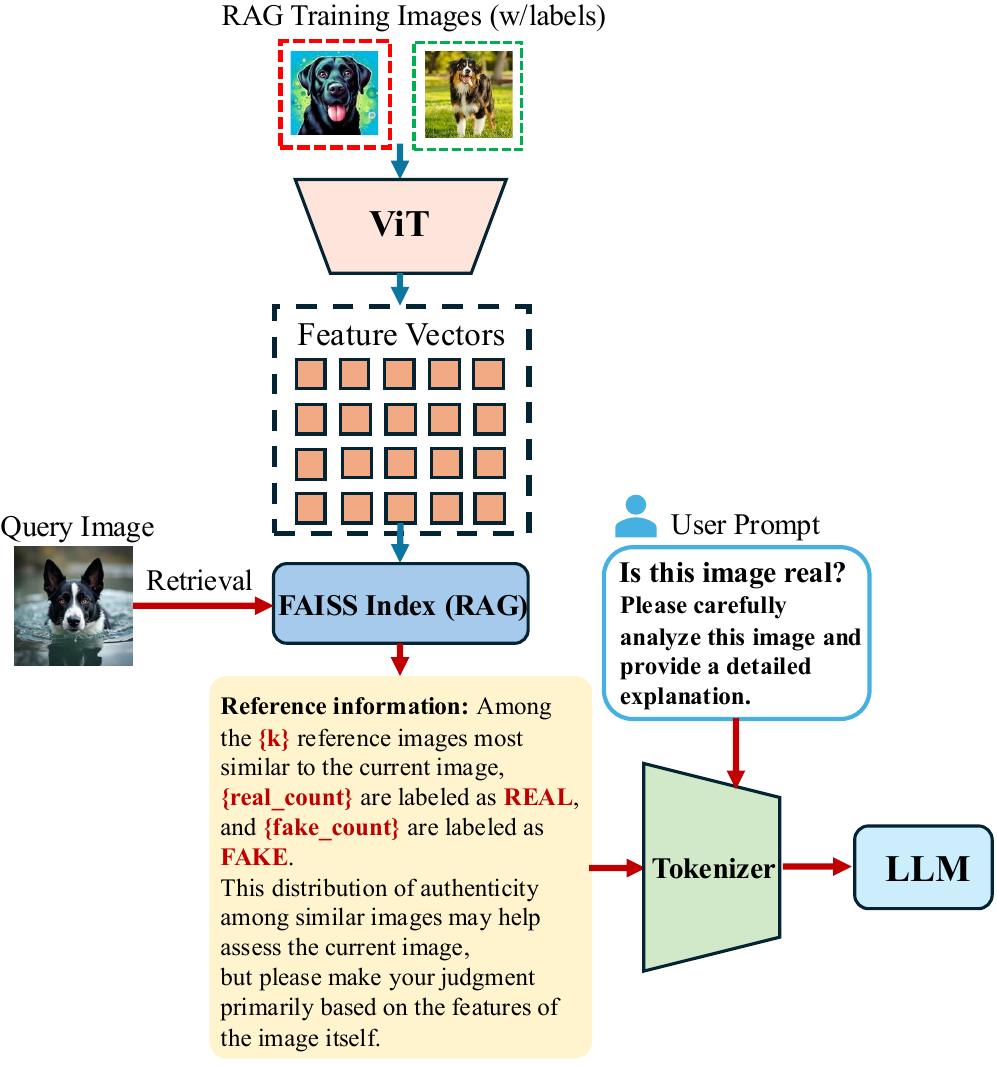}
  \caption{Retrieval-Augmented Generation (RAG) module.}
  \label{fig:rag_module}
\end{figure}

\noindent
\textbf{Retrieval-Augmented Generation Module.}
To leverage prior knowledge from similar examples, we incorporate a Retrieval-Augmented Generation (RAG) module as a preprocessing step before inputs reach the LLM, as discussed in Section~\ref{sec:intro}. Specifically, we construct a vector index comprising all training images using FAISS \cite{douze2025faisslibrary}, an efficient library for fast similarity search over high-dimensional embeddings. Each image is encoded into a fixed-dimensional vector representation using a ViT encoder. We elaborate the steps below.

Given a query image \( x_q \), its visual embedding \( v_q \) is computed and used to retrieve the top-\(k\) most similar images \(\{x_1^r, x_2^r, \dots, x_k^r\}\) from the FAISS index. Each retrieved image has an associated binary label \(y_i^r \in \{\text{REAL}, \text{FAKE}\}\). We aggregate the label distribution to form a retrieval-based statistical summary as:

\textit{"Among the \( k \) retrieved images, \( n_r \) are REAL and \( n_f \) are FAKE."}

This summary is integrated with a fixed task prompt to create the complete input for the LLM. The final reference information is formatted as depicted in Figure~\ref{fig:rag_module}. RAG enables the LLM to reason contextually while minimizing over-reliance on retrieval bias. As the RAG index is constructed once from training data, it incurs no additional supervision or labeling costs.

\noindent
\textbf{Large Language Model.} 
The LLM, with its core parameters frozen and only LoRA adapters updated during training, performs dual-stage reasoning as follows: First, it generates a \texttt{<think>} block containing detailed reasoning steps, analyzing cues such as lighting consistency, shadow sharpness, edge details, or semantic irregularities. Then, it outputs a final \texttt{<answer>} block indicating whether the image is \texttt{REAL} or \texttt{FAKE}. 

Furthermore, we leverage the LLM to produce saliency maps as visual explanations, highlighting regions most critical to the model’s final decision. These maps are derived from attention scores across the ViT’s layers. Specifically, we use the attention rollout method, recursively aggregating attention weights from lower to higher layers. The attention maps between the [CLS] token and patch tokens are fused to generate a coarse-to-fine importance map. Given a sequence of attention matrices \(\{A^{(l)}\}_{l=1}^{L}\), the cumulative attention is computed as:

\begin{equation}
\tilde{A} = A^{(1)} A^{(2)} \cdots A^{(L)}
\end{equation}

The saliency score for each image patch, derived from the attention weight between the [CLS] token and the corresponding patch token, is upsampled to match the original image’s resolution and visualized as a jet colormap overlay, requiring no pixel-level annotations while producing human-interpretable heatmaps that correlate with textual reasoning and highlight forgery artifacts like inconsistent edges, blurred backgrounds, or distorted facial features.

\subsection{Training}
\label{training}

To improve the factual grounding and interpretability of RAIDX, we employ Group Relative Policy Optimization (GRPO) to fine-tune both the ViT-based vision encoder and the LoRA adapters within the LLM. In contrast, the base language model and the retrieval module remain frozen throughout training to preserve pre-trained linguistic and retrieval capabilities. This selective optimization ensures that the model can adapt its visual understanding and explanatory reasoning while avoiding overfitting to the downstream task. Below, we elaborate on the details of GRPO:

\textbf{Reward Definition.} For each query $q$, the model produces a group of outputs $\{o_1, o_2, \ldots, o_G\}$ sampled from the previous policy $\pi_{\theta_{\text{old}}}$. Each output is assigned a reward $r_i$ based on two criteria:
\begin{itemize}
    \item \textbf{Accuracy:} $r_{\text{acc}} = 1$ if the model predicts correctly, otherwise $0$.
    \item \textbf{Format:} $r_{\text{fmt}} = 1$ if the output adheres to the required structure (e.g., using \texttt{<think>} and \texttt{<answer>} blocks), otherwise $0$.
\end{itemize}

The total reward is computed as:

\begin{align}
r_i = r_{\text{acc},i} + r_{\text{fmt},i}
\end{align}

\textbf{Advantage Computation.} To reduce variance and stabilize optimization, we normalize the reward $r_i$ using the mean $\mu_r$ and standard deviation $\sigma_r$ of the group rewards $\{r_1, \ldots, r_G\}$:

\begin{align}
A_i = \frac{r_i - \mu_r}{\sigma_r}, \quad \text{where } \mu_r = \frac{1}{G} \sum_{j=1}^{G} r_j
\end{align}

\textbf{GRPO Objective.} The optimization objective of GRPO is defined as:
\begin{align}
\mathcal{J}_{\text{GRPO}}(\theta) &= \mathbb{E}_{q \sim P(Q)} \Bigg[ \frac{1}{G} \sum_{i=1}^{G} \Bigg( 
\min \left( \frac{\pi_\theta(o_i|q)}{\pi_{\theta_{\text{old}}}(o_i|q)}, \right. \nonumber\\
&\qquad\left. \text{clip}\left(\frac{\pi_\theta(o_i|q)}{\pi_{\theta_{\text{old}}}(o_i|q)}, 1 - \epsilon, 1 + \epsilon \right) \right) A_i \nonumber\\
&\qquad - \beta \, \mathbb{D}_{\text{KL}}(\pi_\theta \parallel \pi_{\text{ref}}) \Bigg) \Bigg]
\end{align}
\noindent where $\epsilon$ and $\beta$ are hyper-parameters, and $A_i$ is the advantage, computed using a group of rewards $\{r_1, r_2, \ldots, r_G\}$ corresponding to the outputs within each group.

The KL-divergence penalty is computed as:
\[
\mathbb{D}_{\text{KL}}(\pi_\theta || \pi_{\text{ref}}) = \frac{\pi_{\text{ref}}(o_i|q)}{\pi_\theta(o_i|q)} - \log \frac{\pi_{\text{ref}}(o_i|q)}{\pi_\theta(o_i|q)} - 1
\]

\noindent Here, $\epsilon$ and $\beta$ are hyperparameters, and $\pi_{\text{ref}}$ refers to a reference policy, such as a pre-trained model or a snapshot from a previous iteration.

%% file: sec/4_experiments.tex
\section{Experiments}
\label{sec:experiments}

In this section, we conduct a comprehensive evaluation of RAIDX, addressing six critical research objectives: (1) systematically validating detection accuracy against standardized benchmarks, (2) assessing cross-model generalization capabilities across unseen generative models, (3) conducting both quantitative and qualitative evaluations on the explanation quality of RAIDX, (4) investigating operational robustness under common real-world perturbations, (5) conducting comprehensive ablation analyses to quantify the impact of RAIDX’s core design choices, and (6) presenting and analyzing both correct and failure cases generated by the RAIDX model to provide deeper insights into its performance.

\subsection{Detection Performance on SID-Set}
\label{sec:sub:performance}

In this experiment, we adopt the SID-Set~\cite{huang2025sidasocialmediaimage} benchmark, which comprises real images and synthetic images generated by a diverse set of foundation models.

Importantly, we emphasize that only the Real and Synthetic subsets of SID-Set are used in both training and evaluation, while the Tampered subset is excluded. This is because our task focuses on whole-image AI-generated detection, whereas the Tampered subset consists of partially manipulated images that fall outside the scope of our detection objective.

We compare RAIDX with a broad set of state-of-the-art baselines, including traditional frequency-domain methods, CNN-based detectors, and recent VLM-based architectures. As metrics, we use Accuracy and F1-score, computed separately on real and fake samples.
\input{tables/sidset_comparison}
As shown in Table~\ref{tab:real_fake_comparison}, RAIDX consistently achieves the highest Accuracy and F1-score across both categories, highlighting its superior detection capability and balanced classification performance. Significantly, RAIDX outperforms the strong baseline SIDA-13B\cite{huang2025sidasocialmediaimage}, published in CVPR 2025, on each evaluation metric, demonstrating its effectiveness.

\subsection{Generalization to Unseen Generative Models}
\label{sec:sub:generalization}

To evaluate RAIDX's zero-shot generalization capability, we adhere to the benchmark protocol of AntifakePrompt~\cite{DBLP:journals/corr/abs-2310-17419}. Specifically, we train RAIDX on its official training set, comprising real images from COCO~\cite{lin2015microsoftcococommonobjects} and Flickr\cite{young-etal-2014-image} alongside synthetic images generated by eight models: Stable Diffusion (v1.4, v1.5, v2.0, v2.1)\cite{stablediffusion-v2-0}, DALLE 2\cite{ramesh2022hierarchicaltextconditionalimagegeneration}, Midjourney v5\cite{midjourneyv5}, IF\cite{deepfloyd-if}, and GLIDE\cite{nichol2022glidephotorealisticimagegeneration}.

For evaluation, we leverage 18 additional datasets from the AntifakePrompt benchmark, all generated by unseen diffusion models such as SDXL\cite{podell2023sdxlimprovinglatentdiffusion}, DALLE-3\cite{dalle3}, DiffusionDB\cite{wang2023diffusiondblargescalepromptgallery}, and GLIDE\cite{nichol2022glidephotorealisticimagegeneration} Stylization. These datasets span diverse styles and domains, ensuring a rigorous test of generalization under distribution shifts.
\input{tables/generalization_table.tex}

As shown in Table~\ref{tab:generalization_comparison}, RAIDX outperforms a wide range of baselines—including frequency-based, CNN-based, and vision-language model (VLM)-based methods such as InstructBLIP and AntifakePrompt—achieving the highest average accuracy across all datasets. In particular, RAIDX demonstrates strong robustness on challenging generative sources such as GLIDE\cite{nichol2022glidephotorealisticimagegeneration}, DALLE-3\cite{dalle3}, and Stylization.

It is noteworthy that although models like ResNet perform well on certain real datasets like COCO\cite{lin2015microsoftcococommonobjects}, their performance collapses under domain shifts, highlighting the importance of robust reasoning and contextual retrieval. The superior performance of RAIDX can be attributed to its integration of RAG-based contextual augmentation and LoRA-enhanced policy reasoning.

\subsection{Explanation Study}

\noindent \textbf{GRPO Study:}
To assess the impact of GRPO on explanation quality, we randomly sampled 100 synthetic images from the SID-Set~\cite{huang2025sidasocialmediaimage}. Each image was processed by two variants of the same base model architecture (Qwen2.5-VL): (1) \textbf{SFT}, trained via supervised fine-tuning, and (2) \textbf{RAIDX}, further optimized with GRPO reinforcement learning. For each image, the models generated an explanation for their prediction. To ensure objective and rigorous evaluation, each output was independently scored by ten domain experts blinded to model identities.
\input{tables/explanation_study}
As shown in Table~\ref{tab:explanation_ci}, RAIDX significantly outperforms SFT in terms of explanation quality, achieving a much higher average score with narrower confidence intervals. These results demonstrate the effectiveness of GRPO in enhancing both the accuracy and coherence of model-generated explanations, surpassing the capabilities of supervised fine-tuning alone.

\noindent \textbf{Comparison with Coarse-grained Methods:}
To further validate the fine-grained reasoning capability of RAIDX, we compare it with \textbf{SIDA-13B}\cite{huang2025sidasocialmediaimage}, a state-of-the-art detector that generates region-aware but coarse textual explanations. While SIDA provides general summaries of visual anomalies, RAIDX produces detailed, evidence-grounded justifications that align with specific visual artifacts, demonstrating superior interpretability in high-fidelity forensics.

\input{tables/explainability_compare_sida}

\begin{figure}[ht]  
  \centering
  \includegraphics[width=\linewidth]{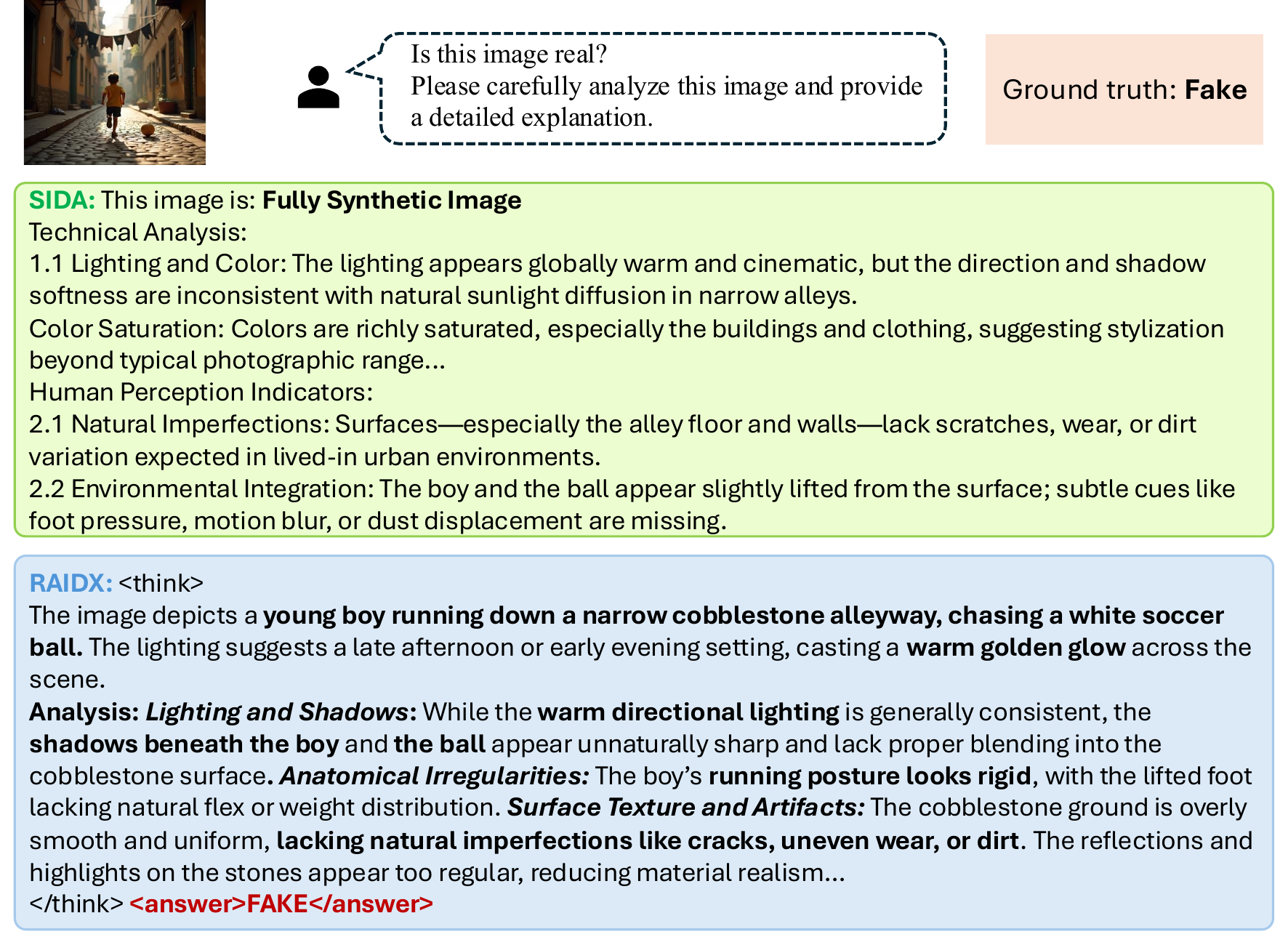}
  \caption{Comparison of explanations from RAIDX and SIDA.}
  \label{fig:explain_visual_compare}
\end{figure}
Table~\ref{tab:explainability_compare_sida} shows that RAIDX significantly outperforms SIDA-13B\cite{huang2025sidasocialmediaimage} in terms of expert-rated explanation quality. Furthermore, Figure~\ref{fig:explain_visual_compare} illustrates a side-by-side visual comparison, where RAIDX highlights precise issues such as eye asymmetry and shadow misalignment, in contrast to SIDA’s broader region-level masks and generic textual feedback.

\subsection{Robustness to Visual Perturbations}
\input{tables/robustness_evaluation}
For robustness evaluation, we apply five commonly observed distortions: JPEG compression (quality 90 and 80), Gaussian blur ($\sigma = 10$), resizing to $224 \times 224$, and brightness reduction with added Gaussian noise. We report performance on two settings: (1) Test1, which applies these perturbations to the SID-Set\cite{huang2025sidasocialmediaimage} test split, and (2) Test2, which applies the same perturbations to 10 unseen datasets from the AntifakePrompt test split—namely SD2, SD3, SDXL\cite{podell2023sdxlimprovinglatentdiffusion}, IF\cite{deepfloyd-if}, DALLE-2\cite{ramesh2022hierarchicaltextconditionalimagegeneration}, DALLE-3\cite{dalle3}, playground v2.5\cite{li2024playgroundv25insightsenhancing}, DiffusionDB\cite{wang2023diffusiondblargescalepromptgallery}, and GLIDE\cite{nichol2022glidephotorealisticimagegeneration}. As shown in Table~\ref{tab:robustness_comparison}, RAIDX was not explicitly trained on distorted images, but it consistently maintains high performance across all degradation types. The accuracy drop compared to clean images is minimal on both test sets, indicating strong generalization capability. These results highlight the model's robustness to real-world noise and compression artifacts, reinforcing its practical utility for deployment in noisy, user-generated content environments.

\subsection{Ablation Studies}
We conduct ablation studies to analyze the individual contributions of the RAG and GRPO modules.

\noindent \textbf{RAG Ablation:} To assess the impact of retrieval-based guidance in RAIDX, we compare three model variants:

\par \textbf{1) No RAG:} The model processes only the raw input image with a generic instruction prompt, without any retrieved context.

\par \textbf{2) Static Prompt:} Instead of dynamic retrieval, a fixed prompt summarizes neighborhood prior statistics in the form: 
\textit{``Reference information: Among the \{k\} reference images most similar to the current image, \{real\_count\} are labeled as REAL, and \{fake\_count\} are labeled as FAKE.''}

\par \textbf{3) RAIDX:} Our full model retrieves top-$k$ similar images using a FAISS index over ViT embeddings. The class distribution of retrieved samples is integrated as contextual guidance for prompt generation.

\par As shown in Table~\ref{tab:rag_ablation_accuracy}, RAIDX outperforms both baselines, improving average accuracy by 1.45\% over ``No RAG'' and 1.48\% over ``Static Prompt''. The gains are especially notable on challenging datasets (e.g., DFDC\cite{dolhansky2020deepfakedetectionchallengedfdc}, GLIDE\cite{nichol2022glidephotorealisticimagegeneration}), confirming the importance of retrieval-grounded instruction over generic or pseudo-context prompts.
\input{tables/ablation_study_on_RAG}

\begin{figure*}[ht]
  \centering
  \includegraphics[width=\linewidth]{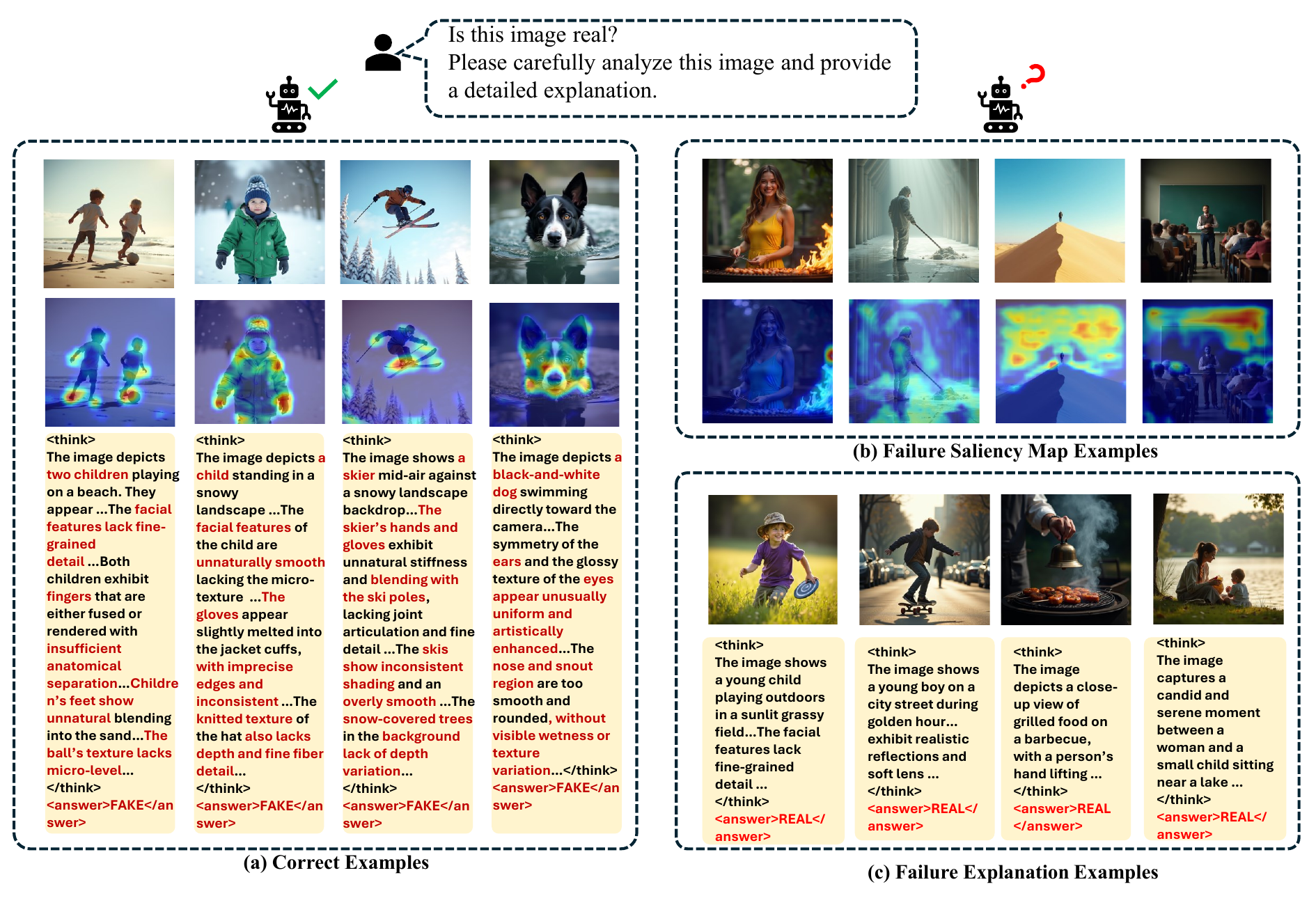}
  \caption{Visualization results from RAIDX, including both correct cases and failure cases.}
  \label{fig:visual_results}
\end{figure*}

\noindent \textbf{GRPO Ablation:}

To rigorously quantify GRPO’s direct impact on classification performance, we performed an ablation on the SID-Set benchmark, holding all data splits, model architecture, and hyperparameters constant. Table~\ref{tab:grpo-ablation} presents the results:

\input{tables/ablation_study_on_GRPO}

Table~\ref{tab:grpo-ablation} shows that adding GRPO to the base model yields a large +30.33\% gain in accuracy, which further improves by +5.62\% when combined with RAG. This step-wise improvement shows that GRPO contributes a substantial and independent gain by jointly optimizing classification accuracy and explanation quality. These results validate GRPO as a critical component of RAIDX’s effectiveness.

\subsection{Case Studies}
In this section, we present a set of representative examples from the SID-Set\cite{huang2025sidasocialmediaimage}, visualized in Figure~\ref{fig:visual_results}. We categorize them into three types: (a) Correct Examples, correctly classifies AI-generated images and produces detailed, visually grounded explanations, (b) Failure Saliency Map Examples, still makes correct predictions, but the saliency heatmaps fail to accurately localize informative regions, and (c) Failure Explanation Examples, illustrates cases where the prediction is incorrect and the explanation lacks visual or semantic specificity.

Overall, RAIDX demonstrates robust performance in both prediction and explanation across most examples. However, the case studies reveal two major failure modes: saliency map drift and failure explanation. These findings reveal two main areas for improvement: improving visual grounding accuracy and encouraging explanations that focus more explicitly on AI-generated artifacts.

%% file: tables/sidset_comparison.tex
\begin{table}[H]
\caption{Performance comparison on the SID-Set.}
\centering
\small
\renewcommand{\arraystretch}{1.2}
\begin{tabular}{llcccc}
\toprule
\multirow{2}{*}{\textbf{Methods}} & \textbf{Year} & \multicolumn{2}{c}{\textbf{Real}} & \multicolumn{2}{c}{\textbf{Fake}} \\
\cmidrule(lr){3-4} \cmidrule(lr){5-6}
& & ACC & F1 & ACC & F1 \\
\midrule
FreDect~\cite{frank2020leveraging} 
& 2020 & 83.7 & 91.1 & 16.8 & 28.8 \\
Gram-Net~\cite{DBLP:conf/cvpr/LiuQT20}
& 2020 & 70.1 & 82.4 & 93.5 & 96.6 \\
CnnSpott~\cite{wang2020cnn}  
& 2021 & 79.8 & 88.7 & 39.5 & 56.6 \\
Fusing~\cite{DBLP:conf/icip/JuJKXNL22}
& 2022 & 85.1 & 92.0 & 34.0 & 50.7 \\
UnivFD~\cite{DBLP:conf/cvpr/OjhaLL23}
& 2023 & 68.0 & 67.4 & 62.1 & 87.5 \\
LGrad~\cite{tan2023learning}
& 2023 & 64.8 & 78.6 & 83.5 & 91.0 \\
LNP~\cite{DBLP:journals/corr/abs-2311-00962} 
& 2023 & 71.2 & 83.2 & 91.8 & 95.7 \\
AntifakePrompt~\cite{DBLP:journals/corr/abs-2310-17419}
& 2024 & 64.8 & 78.6 & 93.8 & 96.8 \\
SIDA-13B~\cite{huang2025sidasocialmediaimage}
& 2025 & 96.7 & 97.3 & 98.7 & 99.3 \\
\midrule
\textbf{RAIDX} & \textbf{2025} & \textbf{98.5} & \textbf{98.9} & \textbf{99.4} & \textbf{99.5} \\
\bottomrule
\end{tabular}
\label{tab:real_fake_comparison}
\end{table}

%% file: tables/generalization_table.tex
\begin{table*}[t]
\centering
\caption{
Accuracy comparison across models and datasets.
\textcolor{green!50!black}{\textbf{Green-shaded rows}} indicate \textbf{Real (non-AI-generated)} datasets, 
while \textcolor{red!75!black}{\textbf{Red-shaded rows}} denote \textbf{AI-generated datasets}.
}

\label{tab:generalization_comparison}
\resizebox{\textwidth}{!}{
\begin{tabular}{lcccccccccccccc}
\toprule
\textbf{Dataset} & \textbf{Ricker2022~\cite{ricker2024detectiondiffusionmodeldeepfakes}} & \textbf{FatFormer~\cite{liu2023forgeryawareadaptivetransformergeneralizable}} & \textbf{Wang2020~\cite{wang2020cnngeneratedimagessurprisinglyeasy}} & \textbf{DIRE~\cite{wang2023dirediffusiongeneratedimagedetection}} & \textbf{LASTED~\cite{wu2023generalizablesyntheticimagedetection}} & \textbf{QAD~\cite{le2023qualityagnosticdeepfakedetectionintramodel}} & \textbf{ResNet~\cite{he2015deepresiduallearningimage}} & \textbf{DE-FAKE~\cite{sha2023defakedetectionattributionfake}} & \textbf{CogVLM~\cite{wang2024cogvlmvisualexpertpretrained}} & \textbf{InstructBLIP~\cite{dai2023instructblipgeneralpurposevisionlanguagemodels}} & \textbf{AntifakePrompt~\cite{DBLP:journals/corr/abs-2310-17419}} & \textbf{RAIDX} \\
\midrule
\rowcolor{green!10}
COCO            & 95.60 & 97.40 & 96.87 & 81.77 & 75.47 & 59.57 & \textbf{99.43} & 85.97 & 98.43 & 97.63 & 92.53 & 92.84 \\
\rowcolor{green!10}
Flickr          & 95.80 & 98.13 & 96.67 & 77.53 & 65.58 & 60.23 & \textbf{99.23} & 84.38 & 99.63 & 97.50 & 91.57 & 94.27 \\
\rowcolor{red!10}
SD2             & 81.10 & 16.83 & 5.23  & 30.47 & 52.53 & 51.00 & 2.50  & 88.07 & 52.47 & 89.57 & 98.33 & \textbf{98.65} \\
\rowcolor{red!10}
SD3             & 88.40 & 21.50 & 8.60  & 98.53 & 79.51 & 46.53 & \textbf{99.83} & 95.17 & 2.10  & 97.60  & 96.17 & 98.13 \\
\rowcolor{red!10}
SDXL            & 81.10 & 30.39 & 1.53  & 19.73 & 77.65 & 41.60 & 0.50  & 72.17 & 32.57 & 96.47 & \textbf{99.17} & 98.77 \\
\rowcolor{red!10}
IF              & 92.65 & 27.73 & 4.93  & 63.17 & 59.89 & 30.97 & 4.40  & 95.20 & 29.03 & 87.90 & 97.10 & \textbf{98.46} \\
\rowcolor{red!10}
DALLE-2         & 52.10 & 76.03 & 3.40  & 61.72 & 59.63 & 15.17 & 12.80 & 61.17 & 60.70 & \textbf{99.27} & 97.27 & 97.91 \\
\rowcolor{red!10}
DALLE-3         & \textbf{95.20} & 43.97 & 8.17  & 36.27 & 51.83 & 9.83  & 2.10  & 71.57 & 6.03  & 67.87  & 80.80 & 87.85 \\
\rowcolor{red!10}
playground v2.5 & 94.40 & 29.83 & 15.73 & 17.73 & 65.42 & 38.73 & 0.20  & 76.07 & 13.37 & 95.43  & \textbf{97.73} & 96.94 \\
\rowcolor{red!10}
DiffusionDB     & 81.20 & 60.50 & 9.64  & 16.40 & 86.48 & 52.07 & 4.69  & 78.10 & 6.05  & 85.40 & \textbf{98.47} & 97.58 \\
\rowcolor{red!10}
SGXL            & \textbf{100.00} & 97.73 & 2.13  & 9.50  & 64.39 & 46.40 & 1.63  & 90.20 & 60.40 & 91.20 & 99.03 & 98.74 \\
\rowcolor{red!10}
GLIDE           & 83.80 & 79.80 & 5.87  & 45.44 & 66.19 & 53.63 & 49.97 & 50.20 & 59.90 & 92.63 & 98.90 & \textbf{98.93} \\
\rowcolor{red!10}
Stylization     & 75.50 & 85.03 & 11.40 & 50.76 & 67.79 & 51.93 & 0.90  & 55.17 & 42.90 & 82.80 & 94.10 & \textbf{95.28} \\
\rowcolor{red!10}
DF              & 14.20 & 5.10  & 0.30  & 3.77  & 58.36 & 97.43 & 34.20 & 77.17 & 13.80 & 67.43 & 95.03 & \textbf{97.93} \\
\rowcolor{red!10}
DFDC            & 46.90 & 1.60  & 0.00  & 60.13 & 70.12 & 90.40 & 14.20 & 48.57 & 9.00  & 85.47 & \textbf{99.83} & 98.84 \\
\rowcolor{red!10}
LaMa            & 64.30 & 67.03 & 7.53  & 13.97 & 60.53 & 42.73 & 1.87  & 23.00 & 5.20  & 42.73 & 39.40 & \textbf{65.98} \\
\rowcolor{red!10}
SD2IP           & 59.10 & 85.07 & 7.23  & 86.40 & 99.56 & 96.30 & \textbf{99.76} & 75.57 & 35.50 & 91.13 & 89.87 & 92.43 \\
\rowcolor{red!10}
SD2SR           & 73.90 & 84.03 & 1.40  & 27.20 & 59.99 & 47.50 & 97.79 & 96.67 & 55.06 & \textbf{99.90} & 99.43 & 92.31 \\
\midrule
\textbf{Average} & 76.40 & 55.98 & 16.23 & 39.48 & 67.83 & 51.78 & 34.78 & 73.58 & 37.90 & 87.10 & 92.48 & \textbf{94.55} \\
\bottomrule
\end{tabular}
}
\end{table*}

%% file: tables/explanation_study.tex
\begin{table}[H]
\centering
\caption{Evaluation on the explainability.}
\label{tab:explanation_ci}
\begin{tabular}{lcc}
\toprule
\textbf{Method} & \textbf{Avg Score} & \textbf{95\% CI} \\
\midrule
SFT      & 22.5  & (20.15, 24.85) \\
\textbf{RAIDX}   & \textbf{82.50} & \textbf{(79.85, 85.15)} \\
\bottomrule
\end{tabular}
\end{table}

%% file: tables/explainability_compare_sida.tex
\begin{table}[H]
\centering
\caption{Explainability Comparison: RAIDX vs. SIDA-13B.}
\label{tab:explainability_compare_sida}
\begin{tabular}{lcc}
\toprule
\textbf{Method} & \textbf{Avg Score} & \textbf{95\% CI} \\
\midrule
SIDA-13B (Coarse-grained)   & 67.15  & (64.89, 69.41) \\
\textbf{RAIDX (Fine-grained)}   & \textbf{82.50} & \textbf{(79.85, 85.15)} \\
\bottomrule
\end{tabular}
\end{table}

%% file: tables/robustness_evaluation.tex
\begin{table}[H]
\centering
\small
\caption{Robustness evaluation of RAIDX on SID-Set\cite{huang2025sidasocialmediaimage} and multiple unseen generative datasets.}
\label{tab:robustness_comparison}
\begin{tabular}{lcccc}
\toprule
\multirow{2}{*}{\textbf{Methods}} & \multicolumn{2}{c}{\textbf{SID-Set (Test1)}} & \multicolumn{2}{c}{\textbf{Average External Sets (Test2)}} \\
\cmidrule(lr){2-3} \cmidrule(lr){4-5}
& ACC & F1 & Avg-ACC & Avg-F1 \\
\midrule
JPEG 90     & 94.21 & 94.02 & 92.35 & 91.97 \\
JPEG 80     & 91.88 & 92.10 & 89.23 & 88.67 \\
Gaussian 10 & 95.72 & 95.41 & 93.68 & 93.21 \\
Resize 224  & 97.33 & 97.21 & 94.83 & 94.65 \\
Bright+Gaussian & 93.45 & 93.16 & 91.28 & 90.84 \\
\midrule
RAIDX & 98.95 & 99.20 & 94.55 & 94.92 \\
\bottomrule
\end{tabular}
\end{table}

%% file: tables/ablation_study_on_RAG.tex
\begin{table}[H]
\centering
\caption{Ablation study of RAG module on detection performance (Accuracy \%).}
\label{tab:rag_ablation_accuracy}
\begin{tabular}{lcccc}
\toprule
\textbf{Dataset} & \textbf{No RAG} & \textbf{Static Prompt} & \textbf{RAIDX} \\
\midrule
SD2           & 97.42 & 97.56 & \textbf{97.65} \\
DALLE-2       & 94.88 & 94.13 & \textbf{97.91} \\
DALLE-3       & 87.11 & 87.27 & \textbf{87.85} \\
LaMa          & 64.83 & 65.48 & \textbf{65.98} \\
GLIDE         & 97.37 & 97.01 & \textbf{98.31} \\
DFDC          & 96.28 & 96.26 & \textbf{98.84} \\
\midrule
\textbf{Average} & 89.64 & 89.61 & \textbf{91.09} \\
\bottomrule
\end{tabular}
\end{table}

%% file: tables/ablation_study_on_GRPO.tex
\begin{table}[H]
  \centering
  \small
  \caption{Ablation of GRPO’s impact on detection accuracy}
  \label{tab:grpo-ablation}
  \resizebox{\columnwidth}{!}{
  \begin{tabular}{lcc}
    \toprule
    \textbf{Model Variant}                & \textbf{Accuracy} & \textbf{ vs Base} \\
    \midrule
    Base                            & 57.12\%           & —                      \\
    Base + GRPO                      & 87.45\%           & +30.33                  \\
    RAIDX (Base + RAG + GRPO)             & 93.07\%           & +35.95                  \\
    \bottomrule
  \end{tabular}
  }
\end{table}

%% file: sec/5_conclusion.tex
\section{Conclusion and Discussion}
\label{sec:conclusion}

In this work, we introduce RAIDX, a novel LLM-based deepfake detection framework that integrates Retrieval-Augmented Generation (RAG) and Group Relative Policy Optimization (GRPO) to simultaneously enhance detection accuracy and decision explainability. By unifying retrieval-augmented exemplar context with reinforcement learning for Chain-of-Thought reasoning, RAIDX combines detection and explanation in a single pipeline. The proposed architecture generates textual rationales and attention-based heatmaps, achieving dual-modality explainability. Extensive experiments on SID-Set\cite{huang2025sidasocialmediaimage} and 18 unseen generative datasets demonstrate RAIDX’s superiority over state-of-the-art methods in accuracy, generalization, and robustness under perturbations. Ablation studies further validate the critical roles of RAG retrieval, GRPO training, and reward design strategies in RAIDX’s performance.

\textbf{Limitations:} As the first paper in the deepfake detection field to employ RAG (Retrieval-Augmented Generation) and GRPO reinforcement learning training strategies, RAIDX has demonstrated promising results. However, as an initial attempt, it still exhibits the following limitations: \textbf{1. Optimization of RAG Implementation:} Current strategies for utilizing RAG leave substantial room for exploration. A critical challenge lies in efficiently integrating unseen data into external knowledge with minimal cost (e.g., generating only 5-10 corresponding synthetic samples per model). This would enable rapid adaptation to future unseen generative models without requiring adjustments to existing model parameters. \textbf{2. Limitations in Tampered Image Detection:} While RAIDX has achieved notable performance in visualizing saliency maps of potential fake regions in fully synthetic images, it currently fails to address cases of image tampering. Enhancing its capability to handle both synthetic content and manipulated authentic media will constitute a key focus of our future research to expand its applicability across diverse scenarios.

\textbf{Future directions:} In addition to the future work outlined in the Limitations part, our planned research will further extend RAIDX to video-based deepfake detection through temporal modeling and develop human-aligned evaluation metrics for assessing explanation quality using human feedback or LLMs.